\pdfoutput=1
\documentclass[11pt]{article}
\usepackage{EMNLP2022}

\usepackage{times}
\usepackage{latexsym}
\usepackage{graphicx}
\usepackage[T1]{fontenc}
\usepackage[utf8]{inputenc}

\usepackage{microtype}

\usepackage{inconsolata}

\usepackage{placeins}

\title{Improving the Generalizability of Text-Based Emotion Detection by Leveraging Transformers with Psycholinguistic Features}

\author{Sourabh Zanwar \\
  RWTH Aachen University \\
  \texttt{sourabh.zanwar@rwth-aachen.de} \\\And
  Daniel Wiechmann \\
  University of Amsterdam \\
  \texttt{d.wiechmann@uva.nl} \\ \AND
  Yu Qiao \\
  RWTH Aachen University \\
  \texttt{yu.qiao@rwth-aachen.de} \\\And
  Elma Kerz \\
  RWTH Aachen University \\
  \texttt{elma.kerz@ifaar.rwth-aachen.de} \\
  }

\date{}

\begin{document}
\maketitle
\begin{abstract}

In recent years, there has been increased interest in building predictive models that harness natural language processing and machine learning techniques to detect emotions from various text sources, including social media posts, micro-blogs or news articles. Yet, deployment of such models in real-world sentiment and emotion applications faces challenges, in particular poor out-of-domain generalizability. This is likely due to domain-specific differences (e.g., topics, communicative goals, and annotation schemes) that make transfer between different models of emotion recognition difficult. In this work we propose approaches for text-based emotion detection that leverage transformer models (BERT and RoBERTa) in combination with Bidirectional Long Short-Term Memory (BiLSTM) networks trained on a comprehensive set of psycholinguistic features. First, we evaluate the performance of our models within-domain on two benchmark datasets: GoEmotion \cite{demszky2020goemotions} and ISEAR \cite{scherer1994evidence}. Second, we conduct transfer learning experiments on six datasets from the Unified Emotion Dataset \cite{BostanKlinger2018} to evaluate their out-of-domain robustness. We find that the proposed hybrid models improve the ability to generalize to out-of-distribution data compared to a standard transformer-based approach. Moreover, we observe that these models perform competitively on in-domain data.

%In recent years, there has been increased interest in building predictive models that harness NLP and ML to detect emotions from various text sources, including social media. Yet,  deployment of such models in real-world sentiment and emotion applications faces challenges, including poor out-of-domain generalizability (and lack of trust in black box models). This is likely due to domain-specific differences (e.g., topics, communicative goals, and annotation schemes) that make transfer across different models of emotion recognition challenging. In this work we propose approaches for text-based emotion detection that leverage transformer models (BERT and RoBERTa) and Bidirectional Long Short-Term Memory (BiLSTM) networks trained on a comprehensive set of psycholinguistic features. First, in within-domain experiments on two benchmark datasets, GoEmotion (Demszky et al. 2020) and ISEAR (Scheer 1994), we find that our models outperform transformer models . Second, in transfer learning experiments on eight emotion classification datasets, we show that our models (can) improve the ability to generalize to out-of-distribution data compared to a transformer-based approach. These results and those of models based solely on psycholinguistic features suggest that enriching/guiding/grounding model predictions with comprehensive psycholinguistic information can improve model performance and generalizability while providing increased model interpretability.

\end{abstract}

\
\section{Introduction}
Emotions are a key factor affecting all human behavior, which includes rational tasks such as reasoning, decision making, and social interaction \cite{parrott2001emotions,loewenstein2003role,lerner2015emotion,bericat2016sociology}. Although emotions seem to be subjective by nature, they appear in objectively derivable ways in texts. Text-based emotion detection (henceforth TBED) is a branch of sentiment analysis that aims to extract textual features to identify associations with various emotions such as anger, fear, joy, sadness, surprise, etc. TBED is  a rapidly developing interdisciplinary field that brings together insights from cognitive psychology, social sciences, computational linguistics, natural language processing (NLP) and machine learning \cite{canales2014emotion,Acheampong2020comparative,alswaidan2020survey,deng2021survey}. TBED has a wide range of real-world applications, from healthcare \cite{cambria2010sentic}, recommendation systems \cite{majumder2019dialoguernn}, empathic chatbot development \cite{casas2021enhancing}, offensive language detection \cite{plaza2021multi}, social data analysis for business intelligence \cite{cambria2013big,soussan2020improved}, and stock market prediction \cite{xing2018natural}.

The differentiation of emotions and their classification into specific groups and categories is a subfield of affective research and has yielded several theories and models \cite{borod2000neuropsychology,scherer2000psychological,cambria2012hourglass,sander2013models,susanto2020hourglass}. The grouping of models for the classification of emotions generally differs according to whether emotions are conceived as discrete/categorical or as dimensional. Categorical models of emotions, like Ekman’s six basic emotions (anger, disgust, fear, joy, sadness, and surprise) \cite{ekman1992there,ekman1999basic}, assume physiologically distinct basic human emotions. Plutchik’s Wheel of Emotion \cite{plutchik1984emotions} is another categorical model that assumes a set of eight discrete emotions expressed in four opposing pairs (joy–sadness, anger–fear, trust–disgust, and anticipation–surprise). Dimensional emotion models, like the Circumplex Model of \citet{russell1980circumplex}, groups affective states into a vector space of valence (corresponding to sentiment/polarity), arousal (corresponding to a degree of calmness or excitement), and dominance (perceived degree of control over a given situation). 

Current approaches to TBED take the advantage of recent advances in NLP and machine learning, with deep learning techniques achieving state-of-the-art performance on benchmark emotion datasets (see \citealt{Acheampong2020comparative} for recent reviews). However there still remains the issue of out-of-domain generalizability of the existing emotion detection models. The way emotions are conveyed in texts may differ from domain to domain, reflecting differences in topics, communicative goals, target audience, etc. This makes the deployment of  such models in real-world sentiment and emotion applications difficult. The importance of this issue has been increasingly recognized in the TBED literature. For example, \citet{BostanKlinger2018} emphasize that “[j]ournalists ideally tend to be objective when writing articles, authors of microblog posts need to focus on brevity”, and that “emotion expressions in tales are more subtle and implicit than, for instance, in blogs''.  To support future transfer learning and domain adaptation work for TBED, the authors constructed a unified, aggregated emotion detection dataset that encompasses different domains and annotation schemes. 

In this work, we contribute to the improvement of the generalizability of emotion detection models as follows: We build hybrid models that combine pre-trained transformer language models with Bidirectional Long Short-Term Memory (BiLSTM) networks trained, to our knowledge, on the most comprehensive set of psycholinguistic features. We evaluate the performance of the proposed models in two ways: First, we conduct within-corpus emotion classification experiments (training on one corpus and testing on the same) on two emotion benchmark datasets, GoEmotion \cite{demszky2020goemotions} and ISEAR \cite{scherer1994evidence}, to show that such hybrid models outperform pre-trained transformer models. Second, we conduct transfer learning experiments on six popular emotion classification datasets of the Unified Emotion Dataset \cite{BostanKlinger2018} to show that our approach improves the generalizability of emotion classification across domains and emotion taxonomies.
The remainder of the paper is organized as follows: In Section 2, we briefly review recent related work on TBED. Then, in Section 3, we present popular benchmark datasets for emotion detection. Section 4 details the extraction of psycholinguistic features using automated text analysis based on a sliding window approach. In Section 5, we describe our emotion detection models, and in Section 6, we present our experiments and discuss the results. Finally, we conclude with possible directions for future work in Section 7. 

\section{Related Work}

In this section, we focus on previous TBED research conducted on two popular benchmark datasets (GoEmotions, ISEAR) to compare the performance of our models with state-of-the-art emotion recognition models, as well as previous attempts to improve generalizability using transfer learning techniques.

Current work on TBED typically utilizes a variety of linguistic features, such as word or character n-grams, affect lexicons, and word embeddings in combination with a supervised classification model \cite[for recent overviews see,][]{sailunaz2018emotion,acheampong2020text,alswaidan2020survey}. While earlier approaches relied on shallow classifiers, such as a naive Bayes, SVM or MaxEnt classifier, later approaches increasingly relied on deep learning models in combination with different word embedding methods. For example, \citet{polignano2019comparison} proposed an emotion detection model based on the use of long short-term memory (LSTM) and convolutional neural network (CNN) mediated through the use of a level of attention in combination with different word embeddings (GloVe, \citealt{pennington2014glove}, and Fast-Text, \citealt{bojanowski2017enriching}). 

In experiments performed on the ISEAR dataset, \citet{dong2022lexicon} proposed a text emotion distribution learning model based on a lexicon-enhanced multi-task convolutional neural network (LMT-CNN) to jointly solve the tasks of text emotion distribution prediction and emotion label classification. The LMT-CNN model is an end-to-end multi-module deep neural network that utilizes semantic information and linguistic knowledge to predict emotion distributions and labels. Based on comparative experiments on nine commonly used emotion datasets, \citet{dong2022lexicon}  showed that the LMT-CNN model can outperform two previously introduced deep-neural-network-based models: TextCNN, a convolutional neural network for text emotion classification \cite{kim-2014-convolutional} and MT-CNN \cite{zhang2018text}, a multi-task convolutional neural network model that simultaneously predicts the distribution of text emotion and the dominant emotion of the text (see Table \ref{table:1} for numerical details on the performance of these models on the datasets used in the present work). In recent years, TBED research has increasingly relied on transformer-based pre-trained language models \cite{Acheampong2020comparative,demszky2020goemotions,cortiz2021exploring}: For example, \citet{Acheampong2020comparative} perform comparative analyses of BERT \cite{devlin-etal-2019-bert}, RoBERTA \cite{liu2019roberta}, DistilBERT \cite{sanh2019distilbert}, and XLNet \cite{yang2019xlnet} for text-based emotion recognition on the ISEAR dataset. 
While all models were found to be efficient in detecting emotions from text, RoBERTa achieved the highest performance with a detection accuracy of 74.31\%. The currently best-performing model on the ISEAR dataset, reaching a micro-average F1 score of 75.2\%, is \citet{park2021dimensional}. In this work a RoBERTa-Large model was finetuned to learn conditional VAD distributions – obtained from the NRC-VAD lexicon \cite{mohammad2018obtaining} – through supervision of categorical labels. The learned VAD distributions were then used to predict the emotion labels for a given sentence.

For the recently introduced GoEmotions dataset, \citet{demszky2020goemotions} already provided a strong baseline for modeling emotion classification by fine-tuning a BERT-base model. Their model achieved an average F1-score of 64\% over an Ekman-style grouping into six coarse categories. \citet{10.1145/3562007.3562051} conducted comparative experiments with additional transformer-based models – BERT, DistilBERT,
RoBERTa, XLNet, and ELECTRA \cite{clark2020electra} – on the GoEmotions dataset. As in the case of ISEAR, the best performance was achieved by RoBERTa, with an F1-score of 49\% on the full GoEmotions taxonomy (28 emotion categories). 

Previous TBED work has also proposed combinations of different approaches. For example, \citet{seol2008emotion} proposed a hybrid model that combines emotion keywords in a sentence using an emotional keyword dictionary with a knowledge-based artificial neural network that uses domain knowledge. To our knowledge, however, almost no TBED research has investigated hybrid models that combine transformer-based models with (psycho)linguistic features (see, however, \citealt{de2021emotional}, for an exception in Dutch). This is surprising, as such an approach has been successfully applied in related areas, for example personality prediction \cite{mehta2020bottom,kerz2022pushing}.

The available research aimed at improving the generalizability of transformer-based models using transfer learning techniques has so far focused on demonstrating that training on a large dataset of one domain, say Reddit comments, can contribute to increasing model accuracy for different target domains, such as tweets and personal narratives. Specifically, using three different finetuning setups – (1) finetuning BERT only on the target dataset, (2) first finetuning BERT on GoEmotions, then perform transfer learning by replacing the final dense layer, and (3) freezing all layers besides the last layer and finetuning on the target dataset –, \citet{demszky2020goemotions} showed that the GoEmotions dataset generalizes well to other domains and different emotion taxonomies in nine datasets from the Unified Emotion Dataset \cite{BostanKlinger2018}.

\section{Datasets}

We conduct experiments on a total of eight datasets. The within-domain experiments are performed on two benchmark corpora: The GoEmotions dataset \cite{demszky2020goemotions} and the International Survey on Emotion Antecedents and Reactions (ISEAR) dataset \cite{scherer1994evidence}. GoEmotions is the largest available manually annotated dataset for emotion prediction. It consists of 58 thousand Reddit comments, labeled by 80 human raters for 27 emotion categories plus a neutral category. While 83\% of the items of the dataset have received a single label, GoEmotions is strictly speaking a multilabel dataset, as raters were free to select multiple emotions. The dataset has been manually reviewed to remove profanity and offensive language towards a particular ethnicity, gender, sexual orientation, or disability. The ISEAR dataset is a widely used benchmark dataset consisting of personal reports on emotional events written by 3000 people from different cultural backgrounds. It was constructed by collecting questionnaires answered by people that reported on their own emotional events. It contains a total of 7,665 sentences labeled with one of seven emotions: joy, fear, anger, sadness, shame, guilt and disgust. The transfer-learning experiments are conducted on six benchmark datasets from Unified Emotion Dataset \cite{BostanKlinger2018} that were chosen based on their diversity in size and domain: (1) The \textbf{AffectiveText} dataset   \cite{strapparava2007semeval} consists of 1,250 news headlines. The annotation schema follows Ekman’s basic emotions, complemented by valence. It is multi-label annotated via expert annotation and emotion categories are assigned a score from 0 to 100. (2) The \textbf{CrowdFlower} dataset consists of 39,740 tweets annotated via crowdsourcing with one label per tweet. The dataset was previously found to be noisy in comparison with other emotion datasets \cite{BostanKlinger2018}. (3) The dataset \textbf{Electoral-Tweets} \cite{mohammad2015sentiment} targets the domain of elections. It consists of over 100,000 responses to two detailed online questionnaires (the questions targeted emotions, purpose, and style in electoral tweets). The tweets are annotated via crowdsourcing. (4) The Stance Sentiment Emotion Corpus \textbf{SSEC} \cite{schuff2017annotation} is an annotation of 4,868 tweets from the SemEval 2016 Twitter stance and sentiment dataset. It is annotated via expert annotation with multiple emotion labels per tweet following Plutchik’s fundamental emotions. (5) The Twitter Emotion Corpus \textbf{TEC} \cite{mohammad2012emotional} consists of 21,011 tweets. The annotation schema corresponds to Ekman’s model of basic emotions. They collected tweets with hashtags corresponding to the six Ekman emotions: \#anger, \#disgust, \#fear, \#happy, \#sadness, and \#surprise, therefore it is distantly single-label annotated. (6) The Emotion-Stimulus dataset \cite{ghazi2015detecting} has 1,549 sentences with their emotion analysed. The set of annotation labels comprises of Ekman's basic emotions with the addition of shame. (7) The ISEAR\textsubscript{UED}  dataset that is part of the Unified Emotion Dataset has 5,477 sentences with single emotion annotations. This dataset is a filtered version of the original ISEAR dataset described above. \citet{BostanKlinger2018} filter and keep the texts with the labels anger, disgust, joy, sadness and fear for the Unified Emotion Dataset.

\section{Sentence-level measurement of psycholinguistic features}

The datasets were automatically analyzed using an automated text analysis (ATA) system that employs a sliding window technique to compute sentence-level measurements (for recent applications of this tool across various domains, see \citet{qiao2020language} for fake news detection, \citet{kerz2021language} for predicting human affective ratings) and \citet{wiechmann2022measuring} for predicting eye-moving patterns during reading). We extracted a set of 435 psycholinguistic features that can be binned into four groups: (1) features of morpho-syntactic complexity (N=19), (2) features of lexical richness, diversity and sophistication (N=77), (3) readability features (N=14), and (4) lexicon features designed to detect sentiment, emotion and/or affect (N=325). Tokenization, sentence splitting, part-of-speech tagging, lemmatization and syntactic PCFG parsing were performed using Stanford CoreNLP \citep{manning2014stanford}.

The group of \textbf{morpho-syntactic complexity features} includes (i) surface features related to the length of production units, such as the average length of clauses and sentences, (ii) features of the type and frequency of embeddings, such as number of dependent clauses per T-Unit or verb phrases per sentence and (iii) the frequency of particular structure types, such as the number of complex nominals per clause. This group also includes (iv) information-theoretic features of morphological and syntactic complexity based on the Deflate algorithm \citep{deutsch1996rfc1951}. 
The group of \textbf{lexical richness, diversity and sophistication features} includes six different subtypes: (i) lexical density features, such as the ratio of the number of lexical (as opposed to grammatical) words to the total number of words in a text, (ii) lexical variation, i.e. the range of vocabulary as manifested in language use, captured by text-size corrected type-token ratio, (iii) lexical sophistication, i.e. the proportion of relatively unusual or advanced words in a text, such as the number of words from the New General Service List \citep{browne2013new}, (iv) psycholinguistic norms of words, such as the average age of acquisition of the word \citep{kuperman2012age} and two recently introduced types of features: (v) word prevalence features that capture the number of people who know the word \citep{brysbaert2019word,johns2020estimating} and (vi)  register-based n-gram frequency features that take into account both frequency rank and the number of word n-grams ($n\in [1,5]$). The latter were derived from the five register subcomponents of the Contemporary Corpus of American English \citep[COCA, 560 million words,][]{davies2008corpus}: spoken, magazine, fiction, news and academic language \citep[see][for details see e.g.]{kerz2020becoming}. The group of  \textbf{readability features} combines a word familiarity variable defined by a prespecified vocabulary resource to estimate semantic difficulty along with a syntactic variable, such as average sentence length. Examples of these measures include the Fry index \citep{fry1968readability} or the SMOG \citep{mclaughlin1969clearing}. 
The group of  \textbf{lexicon-based sentiment/emotion/affect features} was derived from a total of ten lexicons that have been successfully used in personality detection, emotion recognition and sentiment analysis research: (1) The Affective Norms for English Words (ANEW) \citep{bradley1999affective}, (2) the ANEW-Emo lexicons \citep{stevenson2007characterization}, (3) DepecheMood++ \citep{araque2019depechemood++}, (4) the Geneva Affect Label Coder (GALC) \citep{scherer2005emotions}, (5) General Inquirer \citep{stone1966general}, (6) the LIWC dictionary \citep{pennebaker2001linguistic}, (7) the NRC Word-Emotion Association Lexicon \citep{mohammad2013crowdsourcing}, (8) the NRC Valence, Arousal, and Dominance lexicon \citep{mohammad2018obtaining}, (9) SenticNet \citep{cambria2010senticnet}, and (10) the Sentiment140 lexicon \citep{MohammadKZ2013}.

\FloatBarrier

\section{Modeling Approach}
We construct a total of five models: (1) a fine-tuned Bidirectional Encoder Representations from Transformers (BERT) model, (2) a fine-tuned RoBERTA model (Robustly Optimized BERT pre-training Approach), (3) a bidirectional neural network classifiers trained on sentence-level measurements of psycholinguistic features described in Section 3.1,  and (4) and (5) two hybrid models integrating BERT and RoBERTa predictions with the psycholinguistic features. We train all models in a multi-label classification setup. For the within-domain evaluation of the models on the GoEmotions dataset, we follow the  procedure specified in \citet{demszky2020goemotions}: That is, we filtered out emotion labels selected by only a single annotator. The 93\% of the original were randomly split into train (80\%), dev (10\%) and test (10\%) sets. These splits are identical to those used by \citeauthor{demszky2020goemotions}. In the transfer learning setting geared to show that our modeling approach improves generalization across domains and taxonomies, we perform experiments on each of the six emotion benchmark datasets presented in section 3 using four approaches: with/without finetuning on target dataset and with/without the inclusion of the label `neutral'. The performance of these models is evaluated using 5 times repeated 5-fold crossvalidation using a 80/20 split to counter variability due to weight initialization. We report performance metrics averaged over all runs. All models are implemented using PyTorch \cite{NEURIPS2019_9015}. Unless specifically stated otherwise, we use `BCELoss' as our loss function, `AdamW' as optimizer, with learning rate $2\times10^{-5}$ and weight decay of $1\times 10^{-5}$

\subsection{Transformer-based models (BERT, RoBERTa)
} \label{sec:BERT}

We used the pretrained `bert-base-uncased' and `roberta-base' models from the Huggingface Transformers library \cite{wolf2020transformers}. The models consist of 12 Transformer layers with hidden size 768 and 12 attention heads. We run experiments with (1) a linear fully-connected layer for classification as well as with (2) an intermediate bidirectional LSTM layer with 256 hidden units \cite{9078946} (BERT-BiLSTM). The following hyperparameters are used for fine-tuning: a fixed learning rate of $2\times10^{-5}$ is applied and $L2$ regularization of $1\times 10^{-6}$. All models were trained for 8 epochs, with batch size of 4 and maximum sequence length of 512 and dropout of 0.2. We report the results from the best performing models, i.e. RoBERTa-BiLSTM and BERT-BiLSTM.

\begin{figure}
    \centering
    \includegraphics[width = 0.3\textwidth]{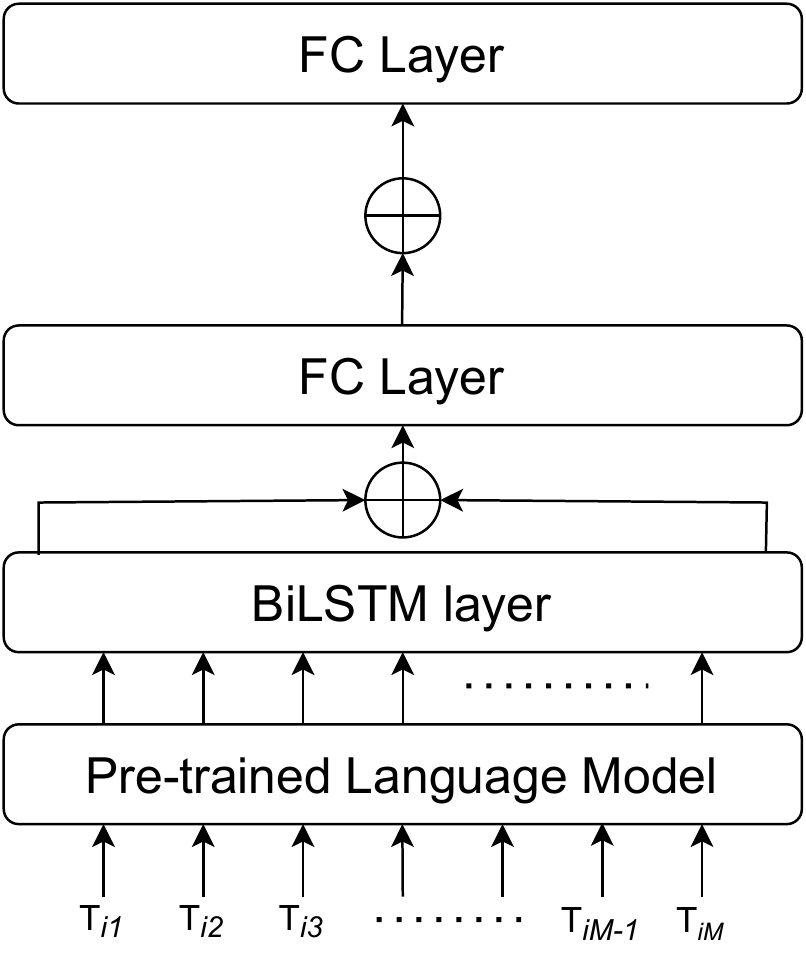}
    \caption{Structure diagram of transformer-based emotion detection models}
    \label{fig:contour_modelfig}
\end{figure}

\subsection{Bidirectional LSTM trained on psycholinguistic features (PsyLing)} \label{sec:contour_model}

\begin{figure}
    \centering
    \includegraphics[width = 0.35\textwidth]{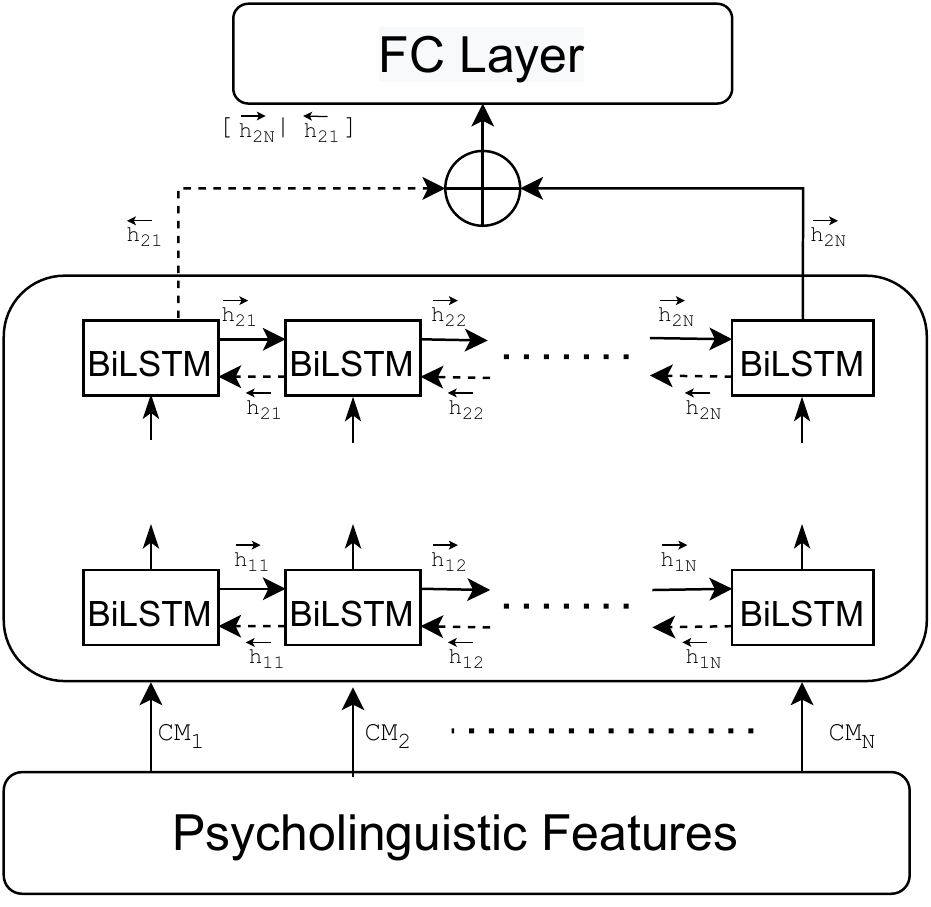}
    \caption{Structure diagram of BiLSTM emotion detection model trained on psycholinguistic features}
    \label{fig:BiLSTM_arch}
\end{figure}

As a model based solely on psycholinguistic features, we constructed a 2-layer bidirectional long short-term model (BiLSTM) with a hidden state dimension of 32, which is depicted in Figure \ref{fig:BiLSTM_arch}.  The input to the model is a sequence  $CM_1^N=(CM_1, CM_2\dots, CM_N)$, where $CM_i$, the output of the ATA-system,  for the $i$th sentence of a document, is a 435 dimensional vector and $N$ is the sequence length. To predict the labels of a sequence, we concatenate the last hidden states of the last layer in forward ($\overrightarrow{h_n}$) and backward directions ($ \overleftarrow{h_n}$). The resulting vector $h_n = [\overrightarrow{h_n}|\overleftarrow{h_n}]$ is then transformed through a 2-layer feedforward neural network, whose activation function is Rectifier Linear Unit (ReLU). The output of this is then passed to a Dense Fully Connected Layer with a dropout of 0.2, and finally fed to a final fully connected layer. The output of this is a $K$ dimensional vector, where $K$ is the number of emotion labels. %More precisely:
% \begin{equation*}
% 	\begin{aligned}
% 	&[\overrightarrow{h_n}, \overleftarrow{h_n}]=\text{BiLSTM}(X) \\
% % 	&h_n=[\overrightarrow{h_n}^T|\overleftarrow{h_n}^T]^T \\
% 	&f = \text{PReLU}(W_{f}h_n + b_{f}) \\
% 	&y=\sigma(W_of+b_o)
% 	\end{aligned}
% \end{equation*}
% where $[\cdot | \cdot]$ is concatenation operator and $\sigma$ is sigmoid function. $\overrightarrow{h_n}, \overleftarrow{h_n}$ are 64 dimensional vectors and their concatenation $h_n=[\overrightarrow{h_n}^T|\overleftarrow{h_n}^T]^T$ is a 128 dimensional vector. $W_f \in \mathbb{R}^{128\times64}$ and $W_o \in \mathbb{R}^{64\times5}$.  Bias terms $b_f$ and $b_o$ are of dimension $64$ and $5$ respectively. The min and max OCLR learning rates are $1\times10^{-5}$ and $1\times10^{-3}$

\subsection{Hybrid models (BERT+PsyLing, RoBERTa+PsyLing)} \label{sec:CNN-CBLF+BERT}
 We assemble the hybrid models by (1) obtaining a set of 256 dimensional vector from the PsyLing model and then (2) concatenating these features along with the output from the pre-trained transformer-based model part. To obtain the output of the pre-trained transformer-based model, the given text is fed to a pre-trained language model, its outputs are passed through a 2-layer BiLSTM with hidden size of 512. This is further passed through a fully connected layer to obtain a 256 dimensional vector. This concatenated vector is then fed into a 2-layer feedforward classifier. To obtain the soft labels (probabilities that a text belongs to the corresponding emotion label), sigmoid was applied to each dimension of the output vector.
 
 %Following Lee et al. \shortcite{lee2021pushing}
% We assemble our hybrid models by (1) obtaining soft labels $C_i$ (probabilities that a text belongs to the corresponding emotion label) from the \note{PsyLing} model by \note{applying sigmoid layer on top of its output logits (True?)} and then (2) interweaving $C_i$ with $K_{im}, m\in[1,M]$ by concatenating $C_i$ with each of $K_{im}$. $K_{im}$ is obtained by feeding the output of Pre-trained Lanaguage model to a \note{describe LSTM layer}, whose output is further transformed by XXXX (3) The concatenated vector is fed into a 2-layer feedforward classifier. %Specifically:

% \begin{equation*}
% 	\begin{aligned}
% 		&f_1 = \text{PReLU}(W_1h_n + b_1)\\
% % 		&f_2 = \left[ f_1^T | g^T\right]^T \\
% 		&f_3 = \text{PReLU}(W_3f_2 + b_3)\\
% 		&y = \sigma(W_of_3 + b_o)\\
% 	\end{aligned}
% \end{equation*}

% where $f_2 = [ f_1^T | g^T]^T$, $W_1\in\mathbb{R}^{64\times32}, W_3\in\mathbb{R}^{37\times32}, W_o\in\mathbb{R}^{32\times5}$ and $b_1, b_3\in\mathbb{R}^{32}, b_o\in\mathbb{R}^5$. The best minimum and maximum OCLR learning rates are found to be $3\times10^{-5}$ and $1\times10^{-3}$

\begin{figure}
    \centering
    \includegraphics[width = 0.5\textwidth]{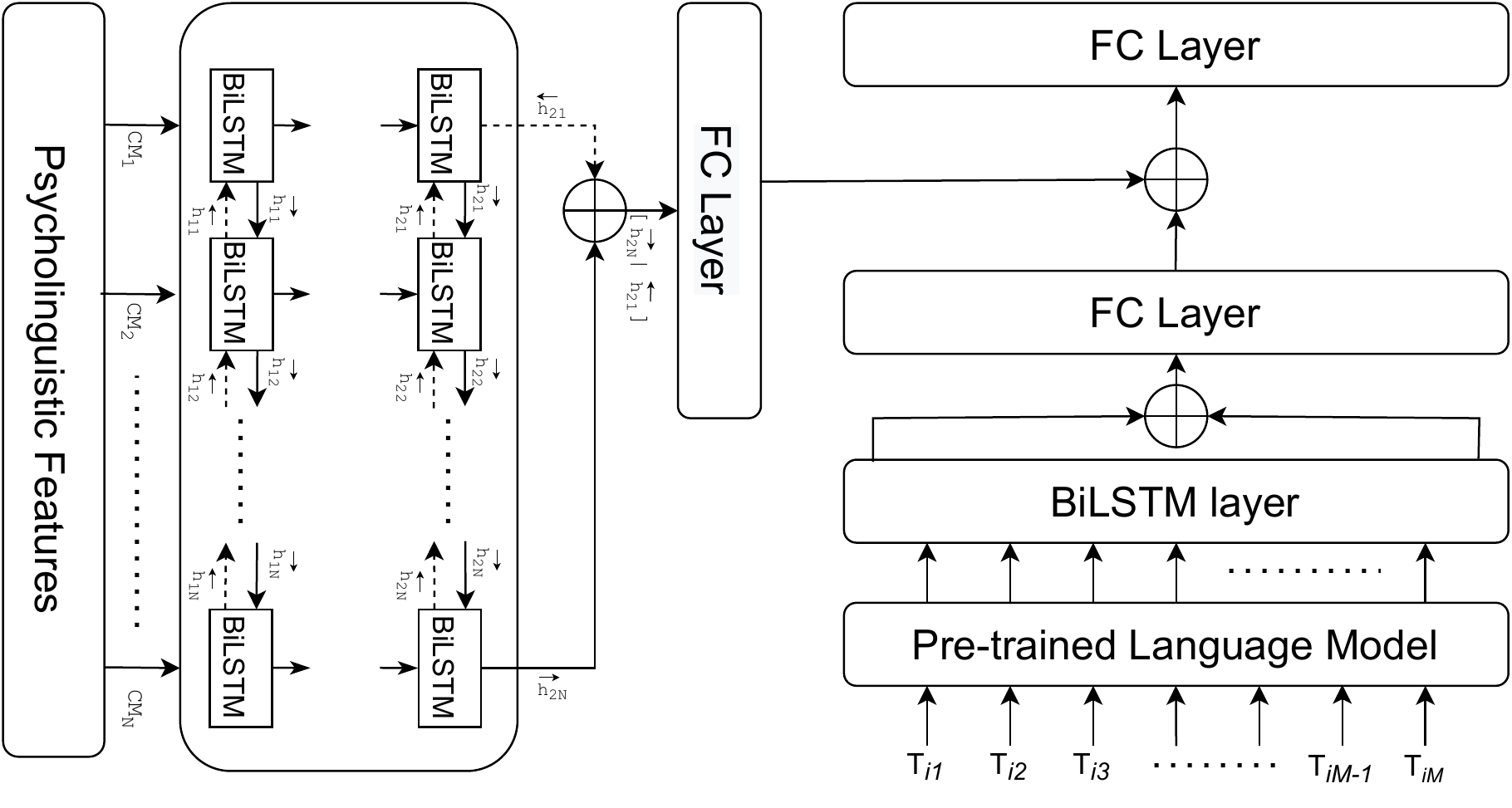}
    \caption{Structure diagram of hybrid emotion detection models}
    \label{fig:hybrid_modelfig}
\end{figure}

\begin{table*}[]
\centering
\setlength{\tabcolsep}{3pt}
\begin{tabular}{llccccccc}
\hline
                &     \multicolumn{7}{c}{\textbf{GoEmotion Dataset}}                                                               \\
                
\hline
Model                            & Anger      & Disgust    &
Sadness    & Surprise   & Fear       & Joy        & Average       \\
\hline
RoBERTa-EMD \cite{park2021dimensional}       & –          & –          & –          & –          & –          & –          & 61.1          \\

%                                & Pre    & 50         & 52         & 56         & 53         & 61         & 77         & 59            \\
%BERT (Demszky et al 2021)       & Rec    & 65         & 53         & 62         & 70         & 76         & 88         & 69            \\
%                                & F1     & 57         & 53         & 59         & 61         & 68         & 82         & 64            \\
%\hline
BERT                                 &   70         &   48         &    64        &   72         &   72         &    90        &    68           \\

RoBERTa                              &    70        &    49        &    63        &     69       &    71        &    90        &       69        \\

PsyLing                              &    50        &    24        &    40        &    40        &     34       &     80       &    45           \\
\hline
\textbf{BERT+PsyLing (ours)}         &    \textbf{71}        &    49        &    \textbf{65}       &     72       &    72        &     91       &    70          \\

\textbf{RoBERTa+PsyLing (ours)}     &    70        &    \textbf{50}        &     \textbf{65}       &    \textbf{74}        &     \textbf{73}       &     \textbf{92}       &     \textbf{71}         \\
\hline
\hline
                           &     \multicolumn{7}{c}{\textbf{ISEAR Dataset}}                                                               \\
                           \hline
%                     &        & \multicolumn{6}{c|}{Emotion}                                                 &               \\
%Model                           & Metric & Anger      & Disgust    &
%Sadness    & Surprise   & Fear       & Joy        & Average       \\    
  % \hline
TextCNN \cite{dong2022lexicon}        & 62.14      & 65.22      & 76.39      & –          & 72.09      & 73.97      & 69.96         \\

MT-CNN \cite{dong2022lexicon}         & 65.68      & 67.63      & 77         & –          & 74.25      & 72.09      & 71.33         \\

LMT-CNN \cite{dong2022lexicon}        & \textbf{66.54}      & \textbf{70.64}      & \textbf{80.68}      & –          & 74.95      & 74.69      & 73.5          \\

RoBERTa-EMD \cite{park2021dimensional}      & \textbf{–} & \textbf{–} & \textbf{–} & \textbf{–} & \textbf{–} & \textbf{–} & \textbf{75.2} \\

BERT                               & 56         & 65         & 71         & -         & 77         & 84         &  71           \\

RoBERTa                          &   60   &   69   &  71    & -   &   72   &   84  &   71    \\

PsyLing                          &  38   &   36   &   48    & -        &   48     &   57   &  45 \\
\hline
\textbf{BERT+PsyLing (ours)}      &  58    &  \textbf{70}   &   70  & -  &  78  &  \textbf{85}  & 72            \\

\textbf{RoBERTa+PsyLing (ours)}     &   \textbf{64}     &   69   &   \textbf{73}    & -   &   \textbf{79}   &   79   & 73  \\
\hline

\end{tabular}
\caption{Results on the two benchmark datasets (GoEmotion (top), ISEAR (bottom)). All scores represent macro-averages of F1 scores(in \%).}
\label{table:1}
\end{table*}

\section{Results}

The models were evaluated using accuracy, precision, recall and F1 scores as the performance metrics. The results of the within-domain classification experiments on the GoEmotion and ISEAR datasets are shown in Table \ref{table:1} (detailed results on all metrics are provided in see Table \ref{table:4} in the appendix). We focus here on the discussion of F1 scores. For both datasets and for both transformer-based models, we find that the proposed hybrid models outperform the standard transformer-based baseline models: Specifically, in the case of the GoEmotions dataset, the hybrid models (BERT+PsyLing, RoBERTa+PsyLing) exhibit an increase in F1 score of +2\% relative to their respective baseline models. In the case of the ISEAR dataset, the RoBERTa+PsyLing model show an increase in F1 score of +2\% relative to RoBERTa, while the BERT+PsyLing model show an increase in F1 score of +1\% relative to BERT. Our hybrid models show improvements in all emotion categories, except for anger, where they are on par with their respective baseline models. These results indicate that integrating transformer-based models with BiLSTM trained on psycholinguistic features can improve emotion classification within two distinct domains: an online domain (Reddit) as well as the domain of reports of personal events. On the GoEmotion dataset, our best-performing hybrid model, RoBERTa+PsyLing, outperforms the previous SOTA model Roberta-EMD \cite{park2021dimensional} by +9.9\% macro-F1. On the ISEAR dataset, both hybrid models outperform two of the three CNNs presented in \citet{dong2022lexicon}, TextCNN and MT-CNN, and are competitive with the lexicon-enhanced multi-task CNN (LMT-CNN). In fact, both hybrid models outperform the LMT-CNN on two of the five emotion categories, with an increase on the joy category of +10.31\% F1 (LMT-CNN vs. BERT-PsyLing) and an increase on the fear category of +4.05\% F1 (LMT-CNN vs. BERT-PsyLing). The results of the comparisons with previous deep-learning TBED models on the two benchmark datasets thus indicate that the proposed approach constitutes a valuable framework for future TBED efforts. 

\begin{table*}[]
\setlength{\tabcolsep}{2pt}
\begin{tabular}{ll|ccccccc}
\hline
                   &  Model                      & TEC & Crowdfl. & ISEAR\textsubscript{UED} & elect-tweet & affect-text & SSEC & emo-stimulus \\
\hline
Train GoEmo     & BERT                   & 29    &   \textbf{23}   & 44      &     26   &   36      &  19    &   53            \\
w/o finetuning        & RoBERTa                & \textbf{31}   &    \textbf{23}  &  44     &   \textbf{29} &  39 & 21     &    56           \\
w/o neutral                   & PsyLing                & 22   &    18  &  25     &  16 &  23 & 11     &   38            \\
                   & BERT+PsyLing     &  \textbf{31}   &   \textbf{23}     &  44     &     27    &    36    &  21    &    56           \\
                   & RoBERTa+PsyLing  & 29   &   \textbf{23}     &  \textbf{47}     &  27 &  \textbf{40} & \textbf{22}     &   \textbf{61}            \\
                   \hline
w/o finetuning     & BERT                   &  20   &   26    &  35   &  23  &  13    &  16    &    41    \\
with neutral       & RoBERTa                & 22    &   27        &  34     & \textbf{25}  &   14     &  \textbf{18}     &  47                 \\
                   & PsyLing                & 16    &    20         &  17     & 13 &    10  &    08   &  23                  \\
                   & BERT+PsyLing    &  21   &   27   &   35  &   24  &   15   &  17    &     45   \\
                   & RoBERTa+PsyLing  & \textbf{23}   &    \textbf{28}     &  \textbf{36}  & \textbf{25}   &    \textbf{16}         &     17        &  \textbf{49}                 \\
                   \hline
with finetuning    & BERT    &   55  &  31   &    63  & 36 & 54  &  \textbf{32} &  92             \\
w/o neutral        & RoBERTa                & \textbf{56}    &    30         &  \textbf{65}     &       34     &      53   &   \textbf{32}   & \textbf{94}              \\
                   & PsyLing                & 34    &     23        &   41    &      32  &     36      & 24 & 46              \\
                   & BERT+PsyLing    & 55 &  32   &  \textbf{65}  &   39  &   \textbf{57}  & \textbf{32} &  \textbf{94}        \\
                   & RoBERTa+PsyLing  & \textbf{56}    &   \textbf{31} & \textbf{65}      &      \textbf{41}    &       \textbf{57}     &  \textbf{32}    &   \textbf{94}            \\
                   \hline
with finetuning    & BERT    & 46 &  33  &  55  &   33  &  44 & 29 &  96  \\
with neutral       & RoBERTa                &  44   & \textbf{34}    & \textbf{56}      &   30      &     46      & 30     &  95             \\
                   & PsyLing                & 24    & 24    & 35      &   28      &     29      & 30     &  53             \\
                   & BERT+PsyLing    & \textbf{47}  &  34  & 55 & 34   &  \textbf{48}  & 31 &  \textbf{97}  \\
                   & RoBERTa+PsyLing & 46    &    \textbf{34} &  \textbf{56}     &  \textbf{34}    &   47   & \textbf{33}    &  96   \\
                   \hline
\end{tabular}
\caption{Results on transfer learning experiments. Values are macro-averaged F1 scores (in \%).}
\label{table:2}
\end{table*}

\begin{table*}[]
\begin{tabular}{l|rrrrrr}
\hline
\multicolumn{1}{c|}{Dataset} & \multicolumn{1}{l}{BERT} & \multicolumn{1}{l}{RoBERTa} & \multicolumn{1}{l}{PsyLing} & \multicolumn{1}{l}{\begin{tabular}[c]{@{}l@{}}BERT +\\  PsyLing\end{tabular}} & \multicolumn{1}{l}{\begin{tabular}[c]{@{}l@{}}RoBERTa + \\ PsyLing\end{tabular}} & \multicolumn{1}{l}{\begin{tabular}[c]{@{}l@{}}\citeauthor{BostanKlinger2018},\\ 2018\end{tabular}} \\ \hline
TEC                          & 63                       & 64                          & 45                          & \textbf{67}                                                                   & 64                                                                               & 48                                                                                    \\
CrowdFlower                  & 46                       & \textbf{47}                 & 41                          & \textbf{47}                                                                   & \textbf{47}                                                                      & 24                                                                                    \\
ISEAR\textsubscript{UED}                        & 76                       & \textbf{78}                 & 49                          & \textbf{78}                                                                   & \textbf{78}                                                                      & 52                                                                                    \\
elect-tweet                  & \textbf{62}              & \textbf{62}                 & 58                          & \textbf{62}                                                                   & \textbf{62}                                                                      & 31                                                                                    \\
affect-text                  & 63                       & 63                          & 48                          & \textbf{67}                                                                   & \textbf{67}                                                                      & 64                                                                                    \\
SSEC                         & 58                       & 60                          & 45                          & 58                                                                            & 60                                                                               & \textbf{67}                                                                           \\
emo-stimulus                    & 94                       & 96                          & 55                          & \textbf{97}                                                                   & \textbf{97}                                                                      & \textbf{97}                                                                           \\ \hline
\end{tabular}
\caption{Comparison of performance with \citet{BostanKlinger2018}. Values are micro-averaged F1 scores (in \%).}
\label{table:3}
\end{table*}

An overview of the results of the out-of-domain experiments is presented in Table \ref{table:2}. Table \ref{table:3} shows comparisons of the results of our best performing model, RoBERTa+PsyLing, in the finetuning setting without the neutral label with the results of maximum entropy classifiers trained on with bag-of-words (BOW) features from \citet{BostanKlinger2018}. The results in Table \ref{table:2} reveal that the RoBERTa+PsyLing hybrid model was the best performing model across all four experimental settings. Performance was generally observed to be highest in the finetuning setting without the neutral label. Importantly, the results in Table \ref{table:2} reveal that the integration of psycholinguistic features matched or improved the performance of the models across all settings, with increases in F1 scores of up to 7\% relative to a standard transformer-based approach. The results in Table \ref{table:3} indicate that our hybrid models pretrained on GoEmotions outperform the results of the baseline models provided by \citet{BostanKlinger2018} on five of the seven emotion datasets (TEC, CrowdFLower, ISEAR\textsubscript{UED}, elect-tweet, and affect text), with increases in performance of up to 31\%. The hybrid models tied the near-perfect performance of the baseline model on the emo-stimulus dataset and fell short only on the SSEC dataset. A possible reason for the relatively low performance of our models on the latter may be due to the fact that the SSEC was rated based on Plutchik’s fundamental emotions.

%Turning to the results of the out-of-domain Similar to the with-in domain experiments, our models show improvements in Out-Of-Domain experiments as well. For Out-of-Domain experiments we use our models on the Unified Emotions Dataset \cite{BostanKlinger2018}.  We either directly infer from the models trained on GoEmotion or finetune the models on the target dataset and then infer.  

\section{Conclusion}

This paper proposed approaches for text-based emotion detection that leverage transformer models in combination with Bidirectional Long Short-Term Memory networks trained on a comprehensive set of psycholinguistic features. The results of transfer learning experiments performed on six out-of-domain emotion datasets demonstrated that the proposed hybrid models can substantially improve model generalizability to out-of-distribution data compared to a standard transformer-based model. Moreover, we found that these models perform competitively on in-domain data. In future work, we intend to extend this line of work to dimensional emotion models as well as to models that jointly solve the tasks of emotion label classification and text emotion distribution prediction.

\section*{Ethical Considerations}

The datasets used in this study may contain biases, are not representative of global diversity and may contain potentially problematic content. Potential biases in the data include: Inherent biases in user base biases, the offensive/vulgar word lists used for data filtering, inherent or unconscious bias in assessment of offensive identity labels. All these likely affect labeling, precision, and recall for a trained model.

%\section*{Acknowledgments}

%The acknowledgments should go immediately before the references. Do not number the acknowledgments section.
%Do not include this section when submitting your paper for review.

\FloatBarrier

\bibliography{acl2020}
\bibliographystyle{acl_natbib}

\clearpage
\FloatBarrier
% \cleardoublepage
\appendix

\onecolumn
% \leftlinenumbers
\section{Appendix}
\label{sec:appendix}

\FloatBarrier
\begin{table*}[h!]
\setlength{\tabcolsep}{3pt}
\caption{Detailed Results on the two benchmark datasets (GoEmotion (top), ISEAR (bottom))}
\begin{tabular}{ll|cccccc|c}
\hline
                &     \multicolumn{8}{c}{GoEmotion Dataset}                                                               \\
                
\hline
%    &        & \multicolumn{6}{c|}{Emotion}                                                 &               \\
Model                           & Metric & Anger      & Disgust    &
Sadness    & Surprise   & Fear       & Joy        & Average       \\
\hline
RoBERTa-EMD (Park et al 2021)   & F1     & –          & –          & –          & –          & –          & –          & 61.1          \\
\hline
%                                & Pre    & 50         & 52         & 56         & 53         & 61         & 77         & 59            \\
%BERT (Demszky et al 2021)       & Rec    & 65         & 53         & 62         & 70         & 76         & 88         & 69            \\
%                                & F1     & 57         & 53         & 59         & 61         & 68         & 82         & 64            \\
%\hline
                                & Pre    & 69         & 38         & 53        &  68         & 68       & 88         &  64             \\
BERT                            & Rec    &  71          &  65          &   80         &  77          &   76         &   91         &    77           \\
                                & F1     &   70         &   48         &    64        &   72         &   72         &    90        &    68           \\
\hline
                                & Pre    &    70        &    62        &    79        &    78        &   71         &    88        &     75          \\
RoBERTa                         & Rec    &    71        &    41        &    53        &     62       &   70         &    93        &       65        \\
                                & F1     &    70        &    49        &    63        &     69       &    71        &    90        &       69        \\
\hline
                                & Pre    &     48       &    28        &    47        &    43        &    42        &     80       &     48          \\
PsyLing                         & Rec    &    53        &    22        &    34        &    38        &    29        &     80       &      43        \\
                                & F1     &    50        &    24        &    40        &    40        &     34       &     80       &    45           \\
\hline
                                & Pre    &    69        &   65         &    68        &    73        &    81        &   90         &      74         \\
\textbf{BERT+PsyLing (ours)}    & Rec    &    71        &    40        &     63       &     69       &    56        &    90        &      65         \\
                                & F1     &    \textbf{71}        &    49        &    \textbf{65}       &     72       &    72        &     91       &    70          \\
\hline
                                & Pre    &    69        &    65        &    68        &    73        &     81       &     90       &      74         \\
\textbf{RoBERTa+PsyLing (ours)} & Rec    &   71         &    40        &     63       &    69        &     56       &     90       &       65        \\
                                & F1     &    70        &    \textbf{50}        &     \textbf{65}       &    \textbf{74}        &     \textbf{73}       &     \textbf{92}       &     \textbf{71}         \\
\hline
\hline
                           &     \multicolumn{8}{c}{ISEAR Dataset}                                                               \\
                           \hline
%                     &        & \multicolumn{6}{c|}{Emotion}                                                 &               \\
%Model                           & Metric & Anger      & Disgust    &
%Sadness    & Surprise   & Fear       & Joy        & Average       \\    
  % \hline
                                & Pre    & 61.36      & 63.5       & 76.64      & –          & 70.67      & 79.3       & 70.29         \\
TextCNN (Dong \& Zeng 2022)     & Rec    & 70.84      & 64.24      & 74.21      & –          & 71.66      & 64.59      & 69.11         \\
                                & F1     & 62.14      & 65.22      & 76.39      & –          & 72.09      & 73.97      & 69.96         \\
\hline
                                & Pre    & 61.31      & 64.68      & 80.27      & –          & 72.16      & 81.13      & 71.91         \\
MT-CNN (Dong \& Zeng 2022)      & Rec    & 71.62      & 64.46      & 77.37      & –          & 73.66      & 69.36      & 71.29         \\
                                & F1     & 65.68      & 67.63      & 77         & –          & 74.25      & 72.09      & 71.33         \\
\hline
                                & Pre    & 62.28      & 66         & 82.07      & –          & 72.5       & 82.15      & 73            \\
LMT-CNN (Dong \& Zeng 2022)     & Rec    & 72.38      & 65.1       & 79.34      & –          & 74.4       & 71.64      & 72.57         \\
                                & F1     & \textbf{66.54}      & \textbf{70.64}      & \textbf{80.68}      & –          & 74.95      & 74.69      & 73.5          \\
\hline
RoBERTa-EMD (Park et al 2021)   & F1     & \textbf{–} & \textbf{–} & \textbf{–} & \textbf{–} & \textbf{–} & \textbf{–} & \textbf{75.2} \\
\hline
                                & Pre    & 51         & 74         & 74         & -         & 83         & 84         & 73            \\
BERT                            & Rec    & 63         & 60         & 69         & -         & 74         & 86         & 70            \\
                                & F1     & 56         & 65         & 71         & -         & 77         & 84         &  71           \\
\hline
                                & Pre    &   58    &  68   &   77   & -     &  93    &  86   &  77  \\
RoBERTa                         & Rec    &    61  &   66  &  64   & -   &   62   &  77  & 66    \\
                                & F1     &   60   &   69   &  71    & -   &   72   &   84  &   71    \\
\hline
                                & Pre    &   26   &   35    &   37   & -         &   46    &  62    &  41   \\
PsyLing                         & Rec    &   62   &   34   &  63   & -       &   48    &    53   &  41  \\
                                & F1     &  38   &   36   &   48    & -        &   48     &   57   &  45 \\
\hline
                                & Pre    &  55  &    73  &  72   & -   &  80  &  84  &    73  \\
\textbf{BERT+PsyLing (ours)}    & Rec    & 62     & 68  &   68  & -  &  77  & 86   &   72  \\
                                & F1     &  58    &  \textbf{70}   &   70  & -  &  78  &  \textbf{85}  & 72            \\
\hline
                                & Pre    &   66    &  72     &   79    & -   &   80    &   80   &   75  \\
\textbf{RoBERTa+PsyLing (ours)} & Rec    &   66    &  66     &   68   & -   &   77   &   77   &   71 \\
                                & F1     &   \textbf{64}     &   69   &   \textbf{73}    & -   &   \textbf{79}   &   79   & 73  \\
\hline

\end{tabular}
\label{table:4}
\end{table*}

\end{document}